\documentclass[10pt,twocolumn,letterpaper]{article}

\usepackage{cvpr}              % To produce the CAMERA-READY version
\usepackage{graphicx}
\usepackage{amsmath}
\usepackage{amssymb}
\usepackage{booktabs}

\usepackage[pagebackref,breaklinks,colorlinks]{hyperref}

\usepackage[numbers]{natbib}

\usepackage[capitalize]{cleveref}
\crefname{section}{Sec.}{Secs.}
\Crefname{section}{Section}{Sections}
\Crefname{table}{Table}{Tables}
\crefname{table}{Tab.}{Tabs.}

\def\confName{CVPR}
\def\confYear{2023}

\begin{document}

\title{DynStatF: An Efficient Feature Fusion Strategy for LiDAR 3D Object Detection}

\author{Yao Rong$^{1*}$, Xiangyu Wei$^{2*}$, Tianwei Lin$^2$, Yueyu Wang$^2$, Enkelejda Kasneci$^3$ \\
%University of T\"ubingen\\
%{\tt\small yao.rong@uni-tuebingen}https://yaorong0921.github.io/homepage/
%\\
%Horizon Robotics\\
%{\tt\small xiangyu.wei@horizon.ai}
%\and
%Horizon Robotics\\
%{\tt\small wzmsltw@gmail.com}
%\and
%\\
%Horizon Robotics\\
%{\tt\small yueyu.wang@horizon.ai}
%\and
%\\
%Technical University of Munich\\
%{\tt\small enkelejda.kasneci@tum.de}
%\and
$^1$University of T\"ubingen, $^2$Horizon Robotics, $^3$University of Munich
}
\maketitle
\def\thefootnote{*}\footnotetext{These authors contributed equally to this work}
%%%%%%%%% ABSTRACT
\begin{abstract}
Augmenting LiDAR input with multiple previous frames provides richer semantic information and thus boosts performance in 3D object detection, However, crowded point clouds in multi-frames can hurt the precise position information due to the motion blur and inaccurate point projection. In this work, we propose a novel feature fusion strategy, DynStaF (\textbf{Dyn}amic-\textbf{Sta}tic \textbf{F}usion), which enhances the rich semantic information provided by the multi-frame (dynamic branch) with the accurate location information from the current single-frame (static branch). 
To effectively extract and aggregate complimentary features, DynStaF contains two modules, Neighborhood Cross Attention (NCA) and Dynamic-Static Interaction (DSI), operating through  a dual pathway architecture. 
NCA takes the features in the static branch as queries and the features in the dynamic branch as keys (values). When computing the attention, we address the sparsity of point clouds and take only neighborhood positions into consideration. NCA fuses two features at different feature map scales, followed by DSI providing the comprehensive interaction. To analyze our proposed strategy DynStaF, we conduct extensive experiments on the nuScenes dataset. On the test set, DynStaF increases the performance of PointPillars in NDS by a large margin from 57.7\% to 61.6\%. When combined with CenterPoint, our framework achieves 61.0\% mAP and 67.7\% NDS, leading to state-of-the-art performance without bells and whistles.

%With the , PointPillars is improved 
%on the nuScenes dataset with two popular frameworks, PointPillars and CenterPoint  we improve the popular frameworks PointPillars by a large margin from 57.3\% to 61.4\%, and CenterPoint by 1.5\% on the nuScenes validation dataset, which leads to the competitive state-of-the-art performance without bells and whistles. % on the NuScenes test set in LiDAR 3D object detection with xx\%.
\end{abstract}

%%%%%%%%% BODY TEXT
\section{Introduction}
\label{sec:intro}
%Background of Lidar in 3D object detection, some popular methods: pillarbased and voxelbased. 
LiDAR sensor is widely used for 3D object detection in the context of autonomous driving because of its high precision in depth information. Recent methods based on LiDAR information utilizing Bird's Eye View (BEV) can be mainly categorized into two groups: voxel-based (VoxelNet proposed by \citet{zhou2018voxelnet}) and pillar-based (PointPillar proposed by \citet{lang2019pointpillars}). The former group \cite{zhou2018voxelnet,deng2022vista,ye2022lidarmultinet, yan2018second} first divides points in the space into equally distributed voxels, and obtain features with several 3D convectional layers. The latter one \cite{wang2020pillar,chen2021polarstream,lang2019pointpillars,wang2020infofocus} converts 3D points into pseudo images by generating pillars at each position in a 2D image, whose size on the vertical-axis being equal to the whole available space and thus is able to directly acquire feature representations using 2D convolution. 
%\textcolor{red}{Since points are spare in the space, limited numbers of pillars and points in a pillar are set to hold the sparsity.} 
Without feature compression on the vertical-axis, voxel-based methods yield higher performance, while pillar-based models are more efficient in the computation and preferred in real-time applications. 

\begin{figure}[!t]
    \centering
  \begin{subfigure}[b]{0.45\linewidth}
    \includegraphics[width=\textwidth]{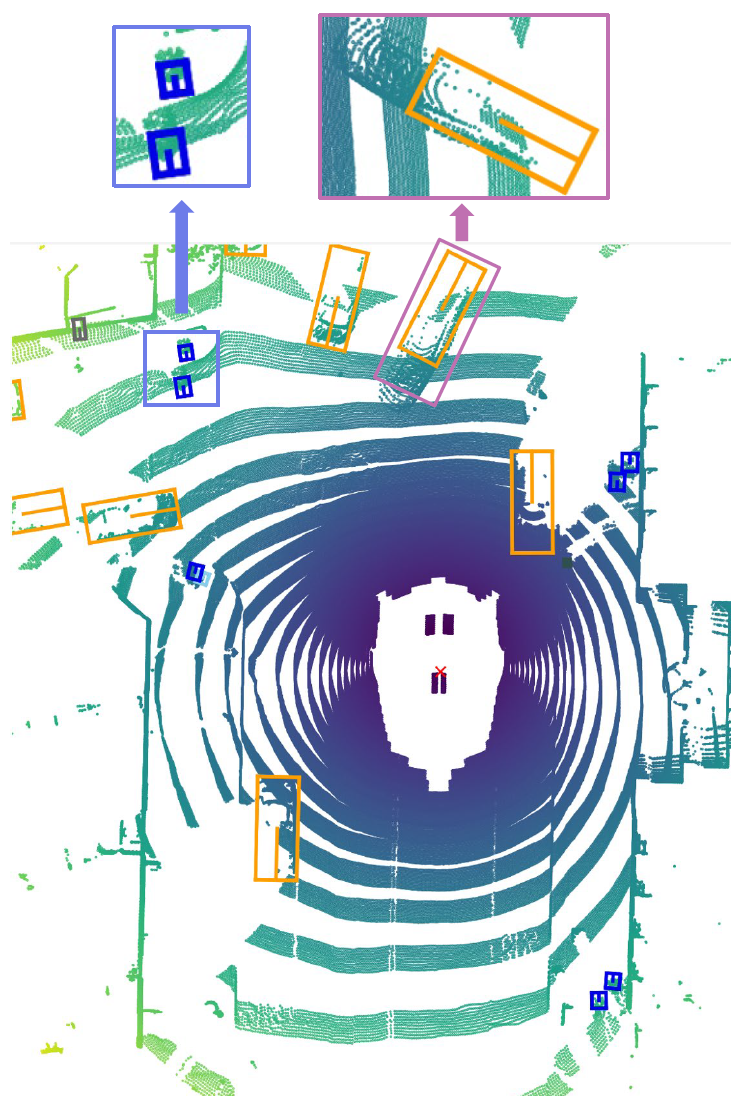}
    \caption{Multi-frame input.}
    \label{fig:f1}
  \end{subfigure}
  \begin{subfigure}[b]{0.45\linewidth}
    \includegraphics[width=\textwidth]{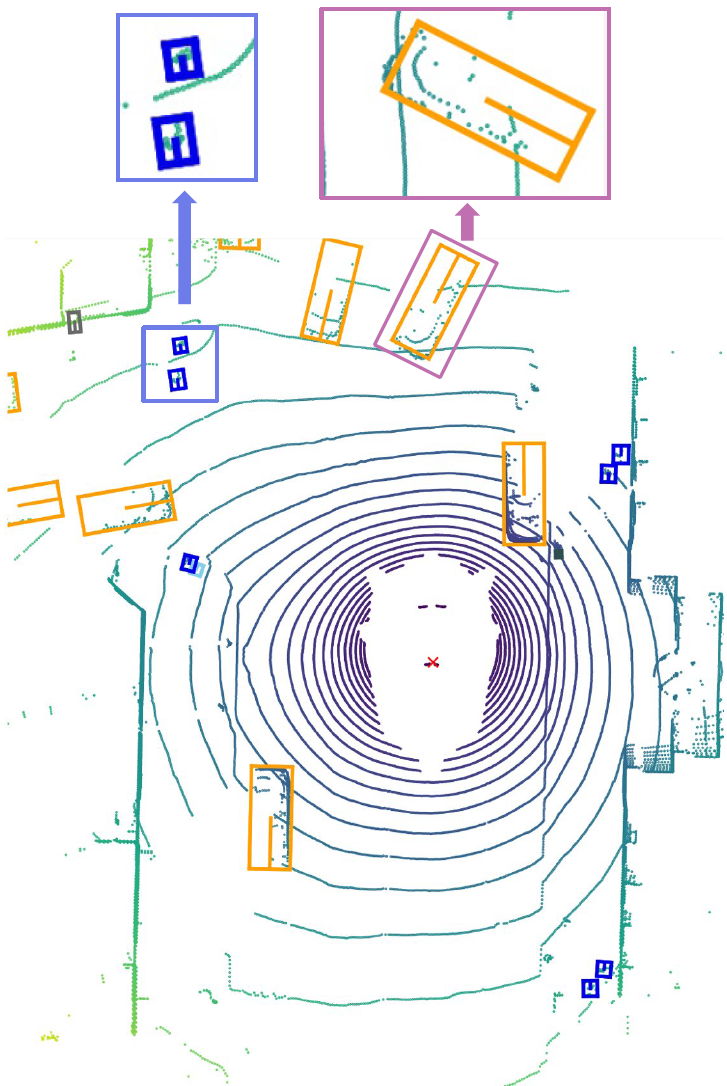}
    \caption{Single-frame input.}
    \label{fig:f2}
  \end{subfigure}
  \caption{Visualization of multi-frame (10 sweeps) and single-frame input. Bounding boxes are ground-truth objects on nuScenes. Zoom-in images above demonstrate a clear view.}
  \label{fig:teaser figure}
\end{figure}

%Advantage and disadvantage of multi-frame and single-frame.
LiDAR point cloud contains precise geometric shapes and exact positions of objects, but it suffers from the irregular point density: points are dense in the area closed to the LiDAR sensor while very sparse far away. Detecting objects with fewer points is very difficult. Suggested by \cite{caesar2020nuscenes,li2022unifying}, using multiple LiDAR sweeps (frames) provides richer point cloud information to eliminate the unclarity due to the sparsity of the points. Points from multiple sweeps are aggregated directly and distinguished by expanding input data with relative timestamp information as an extra dimension, which also enhances the network with valuable temporal information.
%As a result, multi-frame data yields a positive impact on the detection performance while introducing little additional computational cost.
%Nevertheless, it is challenging to keep the odometry measurement precise all the time in practice, which brings inaccurate previous points into the current frame. Moreover, motion blur is inevitable for dynamic objects, especially in busy traffics. 
\Cref{fig:teaser figure} illustrates the difference between the multi-frame and single-frame input. In this scene, there are several vehicles and pedestrians in front of the car. It is easy to determine the location of objects based on the unambiguous points on their surface with a single sweep. However, after accumulating ten sweeps of point clouds, motion blur is observed around vehicles and pedestrians such that edges of moving objects become obscure to be accurately recognized (see zoom-in images), leading to confusion about concrete positions. %\textcolor{red}{For instance, motion blur can be seen for the vehicle in the front and the pedestrian on the upper right corner, resulting in confusion for localization.} 
%On the other hand, multi sweeps enhance the overall shape of objects, for example, two vehicles on the left side have a clearer shape benefiting the detection. 
%For instance, it is challenging for a detector to recognize each individual from a point cloud of multiple moving pedestrians closed to each other. 
In a word, multi-frame LiDAR input boosts recognition performance by augmenting the input with meaningful motion characteristics (\textit{dynamic} information), but suppresses the advantage of a single frame in exact object localization capability (\textit{static} information). 
However, existing works commonly employs only multi-frame point cloud data as input, such as \cite{deng2022vista, huang2022rethinking, zhu2019class, chen2020object} conducting experiments on the large-scale outdoor dataset nuScenes \cite{caesar2020nuscenes}. Based on the observations above, we propose a novel unified framework named DynStaF, standing for Dynamic-Static Fusion, to bridge the current research gap by fusing the rich semantic information provided by the multi-frame input with the accurate location information from the single-frame data effectively. To the best of our knowledge, DynStaF is the first attempt to deploy a two-stream architecture for extracting and fusing features from multi-frame and single-frame LiDAR input.  %\textcolor{red}{Specifically, we design a novel unified strategy named DynStaF standing for Dynamic-Static Fusion.}
%To address the feature interaction within the network, several strategies are explored in this paper. 

DynStaF deploys a dual pathway architecture to operate on BEV features from both input types concurrently across the 2D backbone. To address the feature interaction, we introduce two fusion modules, Neighborhood Cross Attention (NCA) and Dynamic-Static Interaction (DSI) performing feature fusion between two branches at different levels. %\textcolor{blue}{more reasonably}. 
%which is proved to be good at build connections among related features %is employed by us to perform cross attention in feature fusion. 
%Concretely, NCA is a transformer-based module, which generates cross attention between two pathways by regarding features from the single-frame branch as queries and features from the multi-frame branch as keys.
An attention mechanism is adapted to produce the cross attention, but a vanilla cross attention module computes the attention matrix globally. LiDAR BEV feature maps are sparse where correlative features are distributed locally. Considering this characteristic of BEV features, we do not need to compute global attention, as it does not bring significant benefits but introduce an overhead of computation. Thus, we choose to conduct cross-attention limited to the neighborhood area. Concretely, NCA regards features from the single-frame branch as queries and obtains keys and values in the neighborhood of queries from the multi-frame feature map.
%the cross attention is computed to the neighborhood area of each query, since the correlative features should locate in the similar position in both feature maps.
%Moreover, the feature map scattered by non-empty voxels or pillars after the point cloud encoder is relatively sparse. Therefore, computing attention globally does not bring significant benefits but introducing an overhead of computation.
%\textcolor{red}{we first deploy the transformer-based NCA module to achieve the cross attention by using the features from the single-frame branch as queries and the features from the multi-frame branch as keys.} \textcolor{blue}{With the help of NCA, rich semantic information can be fused into the single-frame features to provide enhanced predictions.} 
%The obtained cross attention is further applied on single-frame features to increase the precision of the detected object position, while containing rich semantic features. 
After several blocks in the backbone, the feature maps become dense. At this stage, we utilize the CNN-based DSI module which conducts comprehensive interaction at each pixel position. The fused feature contains rich semantic context and accurate position information that promotes the detection precision.
%Capturing significant features around each single query, expensive computational cost in vanilla transformer block is saved. To further improve the network efficiency, we design a specific operator for point cloud, which only performs attention on non-empty grids, for only a few sparse regions are occupied in point cloud.

% our contribution
In summary, our work has three main contributions:
%The contributions in this paper can be summarized as follows: 
\begin{itemize}
\item We propose a novel feature fusion strategy termed as DynStaF which has a dual pathway architecture to efficiently fuse the complementary information from the multi-frame and single-frame LiDAR input. 

\item %Considering the characteristics of BEV feature maps from dynamic and static branches, we propose two modules for fusion at different levels. 
Taking into account the specific features of BEV feature maps extracted from dynamic and static branches, we introduce two modules designed for fusion at distinct levels.
Neighborhood Cross Attention module is designed for sparse feature maps, while Dynamic-Static Interaction module for dense feature maps.
\item We conduct extensive experiments to analyze and benchmark our methods on the challenging dataset nuScenes. DynStaF boosts the performance of PointPillars \cite{lang2019pointpillars} significantly on the nuScenes test set by 3.9\% in NDS and 5.9\% in mAP. When using CenterPoint \cite{yin2021center} as the backbone, our framework achieves 67.7\% and 61.0\% in NDS and mAP, respectively, surpassing other state-of-the-art methods without bells and whistles. 
\end{itemize}

%As DynStaF does not change the multi-frame branch settings and only contains 2D convolutional blocks, it can be easily plugged into any advanced object detector without introducing heavy computation overhead.

%  On the popular framework, , the performance on the challenging Nuscenes dataset is improved , . Our method on a state-of-the-art detector, CenterPoint, boosts the performance by 1.5\%, which validates the effectiveness of our algorithm.

\section{Related Work}
\paragraph{3D object detection with LiDAR.} The task of 3D object detection based on LiDAR in autonomous driving recently is to detect traffic participants and place 3D bounding boxes around them. Along with the detection, classes as well as attributes of objects (e.g., moving or parked) or other information are estimated \cite{caesar2020nuscenes,Sun_2020_CVPR}. Recent algorithms for LiDAR 3D object detection are all based on BEV feature maps. VoxelNet \cite{zhou2018voxelnet} turns point clouds into voxels and apply first 3D CNNs to encode the voxel features and then 2D convolutions to accomplish the detection. Besides, SECOND is another popular framework deploying 3D convolution operations \cite{yan2018second}. \citet{lang2019pointpillars} propose to encode the point clouds to pillar vectors and then project these pillars onto the 2D BEV space. Frameworks based on PointPillars only need 2D convolutional layers to process the point BEV feature maps for 3D object detection \cite{wang2020pillar,chen2021polarstream,lang2019pointpillars,wang2020infofocus}. To further boost the performance of detectors, CenterPoint \cite{yin2021center} proposes a new detection head, which first detects object centers of and then computes other attributes such as bounding box sizes, followed by refining these estimates in the second phase. It turns to be effective when combining with a 3D backbone such as VoxelNet, which is widely used as a state-of-the-art framework. In this work, we use PointPillars and CenterPoint as our 2D and 3D backbones to show the effectiveness and the compatibility of our DynStaF.

\paragraph{Feature fusion strategy.} Feature fusion can boost the performance in 3D LiDAR object detection. Multi-modality fusion is one popular strategy, for example, \cite{bai2022transfusion, liu2022bevfusion, li2022unifying}
propose frameworks that fuse camera and LiDAR data, where the performance is stronger than using a single modality. Another group of feature fusion works does not require multiple sensors, for instance, \citet{deng2022vista} fuse BEV features with RV (range view) features, as RV provides dense features while BEV features are sparse but not overlapped. Combining two views improves the performance as it gives comprehensive spatial context. HVPR (Hybrid Voxel-Point Representation) \cite{noh2021hvpr} utilizes voxel-based features and point-based features as used in PointNet++ \cite{qi2017pointnet++}. In this way, voxel features, which are effective to be extracted, are integrated with more accurate 3D structures from point streams. As geometric information gets lost when projecting to the 2D BEV space, MDRNet \cite{huang2022rethinking} enriches the BEV features with voxel features to keep the geometry information. In our project, we propose a novel feature fusion strategy based on the characteristics of multi-frame and single-frame LiDAR input, which keeps features in both branches interacting across the whole BEV feature processing. Our strategy is trained end-to-end on the 2D BEV space, which can be directly applied to any state-of-the-art architectures to boost the performance. 
%The feature fusion takes place only once before the detection in many previous feature fusion methods such as \cite{deng2022vista,}, however, our strategy keeps features in both branches interacting with each other across the whole BEV feature processing, which enhances the spatio-temporal information in a comprehensive way. 
% feature fusion strategy: vista, MSR (voxel branches and BEV branches, Rethinking Dimensionality Reduction.. ), HVPR (Hvpr: Hybrid voxel-point.., cleverly keeps the efficiency of pillar-based detection while implicitly leveraging the voxel-based feature learning regime for better performance), lidar-camera fusion (TransFusion: Robust ..., bevfusion it has more papers ) 
%MotionNet proposed by \citet{wu2020motionnet} extracts spatio-temporal feature in a hierarchical fashion while brings significant extra computational cost due to repetitive voxelization and following voxel feature extraction process.
\vspace{-10pt}
\paragraph{Transformer attention mechanism.}
Thanks to the attention mechanism, the transformer architecture is powerful in fusing features from different source or modalities for 3D object detection or other tasks in autonomous driving \cite{deng2022vista, bai2022transfusion,li2022unifying,li2022bevformer,zhou2022centerformer}. In \cite{deng2022vista}, the authors use the BEV features as queries and RV features as keys and values to conduct the cross attention between the two views. 
TransFusion \cite{bai2022transfusion} designs a new detection head based on transformer. In the first stage, a sparse set of object queries from LiDAR BEV features are used to get the initial bounding boxes; In the second stage, another transformer layer is deployed to obtain the cross attention from camera images and LiDAR data. Centerformer \cite{zhou2022centerformer} enhances the center-based object detection by using the center candidates as queries in a DETR-style (DEtection TRansformer \cite{carion2020end}) transformer. Moreover, cross attention between current frame and previous frames are extracted using a deformable DETR \cite{zhu2020deformable}. \citet{li2022bevformer} also deploy deformable DETR to gain the temporal and spatial attention among multi-camera images for the object detection. 
Different from previous work, our method adopts the neighborhood attention mechanism for transformers \cite{hassani2022neighborhood} to get the cross attention between multi-frame and single-frame LiDAR input. As the BEV features are sparse, the object should be at a similar spatial location in both input and thus focusing on neighborhood produces high-quality fusion, which is verified by our experiment results.

\begin{figure*}[t]
    \centering
    \includegraphics[width=\linewidth]{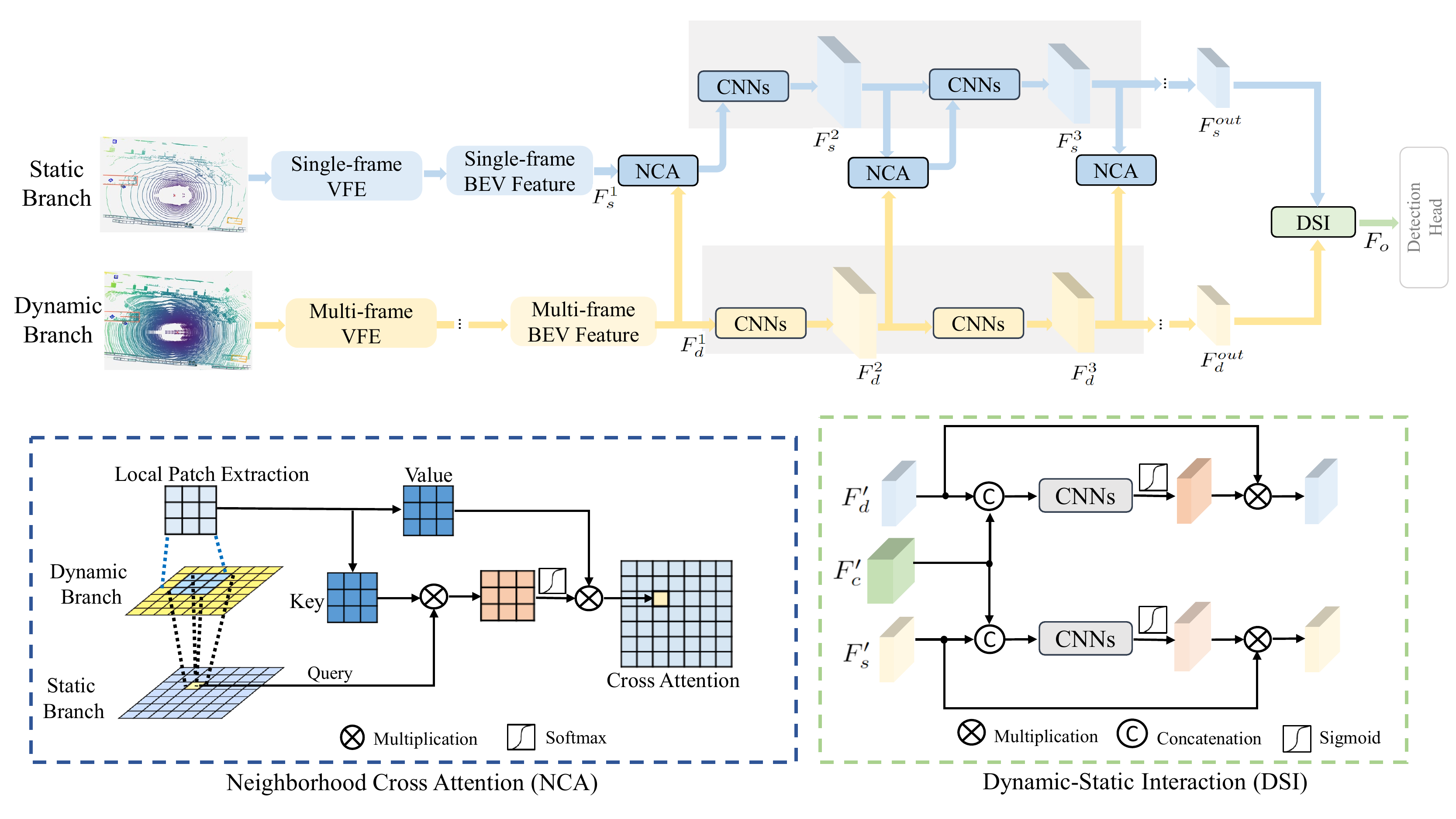}
  \caption{\textbf{Upper}: Overall Architecture of DynStaF. 2D convolutional backbone is highlighted with the gray color. \textbf{Bottom}: The overview of two core fusion modules: Neighborhood Cross Attention (NCA) and Dynamic-Static Interaction (DSI). Channel dimension in NCA is omitted for a clear view.}
  \label{fig:archi}
\end{figure*}

\section{Method}
% Overall architecture. 
% Mathematical expression.
Most recent LiDAR-based 3D detection approaches aggregate raw point clouds from a sequence of LiDAR point clouds and use the (relative) timestamp as an additional feature dimension to improve detection performance. This setting is effective in compensating the sparsity of point clouds with a single frame as input for 3D object detection. As discussed in \Cref{sec:intro}, point clouds from previous frames will bring ambiguity in localization especially for moving objects in crowded scenarios. To mitigate this adverse impact, we propose to deploy cross attention to efficiently fuse spatio-temporal semantic features from input sequence with the accurate localization information from the current frame. A dual pathway architecture is designed to process current and aggregated point clouds separately, where the extracted features are fused progressively. Our framework is termed as ``DynStaF", and we refer the multi-frame branch as ``Dynamic Branch" and the single-frame branch as ``Static Branch" to highlight the rich motion information and accurate location information in each branch. 

% However, mixture of point clouds from different frames will result in some unclarity in bounding box localization and confusion in crowd scenarios, such as busy crossroads. We propose a novel feature fusion strategy termed DynStaF, which takes point cloud sequences as well as current frame data as input, such that both temporal semantic feature and exact positional feature are fused to greatly improve the detection results. We design a dual-path network architecture, which extracts multi-frame feature and single-frame feature in parallel. Moreover, to fuse these features effectively, DynStaF employs transformer-based cross attention. Instead of expensive global attention, we restrict the receptive field of each token to the neighbourhood of itself, which takes full advantage of the sparsity of point cloud features in BEV and reduces the computational cost of vanilla transformer architecture significantly. 

\subsection{Overall Architecture}
% Architecture of the single frame branch.
% multi levels where fusion happens. (Illustration).
We follow general 3D LiDAR object detector settings without requiring any extra input information. Popular detectors such as pillar-based frameworks project point cloud into BEV feature space after voxelization (Voxel Feature Encoding), while voxel-based frameworks usually process the voxels with a 3D backbone additionally. All popular pillar-based/voxel-based architectures can be straightly deployed as the dynamic branch in DynStaF.
Complicated 3D backbones may be used to process voxels to obtain BEV features such as in CenterPoint \cite{yin2021center}. However, the static branch is designed to be light-weighted and only needs VFE to encode the BEV features, i.e., no 3D backbone is needed in the static branch. DynStaF operates on BEV features. The projected BEV feature map of dynamic branch is denoted as $F_{d} \in \mathbb{R}^{C_{d} \times W_{d} \times H_{d}}$ and $F_{s} \in \mathbb{R}^{C_{s} \times W_{s} \times H_{s}}$ for static branch, where $C$, $W$ and $H$ refer to the number of channel, width and height of the generated BEV maps, respectively. Before starting the feature fusion, $F_{s}$ is processed with extra convolutional layers to reach the same dimension as $F_d$ if their dimensions vary. Feature fusion between two branches occurs regressively using NCA in our DynStaF, as illustrate in \Cref{fig:archi} (Upper).

In the $l$-th fusion block ($l \in \{1, 2, ...,  N\}$), given the input dynamic branch feature $F_{d}^l$ and static branch feature $F_{s}^l$, the output can be formulated as:
% Denoting the fusion operation as $\mathcal{A}^i(\cdot)$, 
\begin{equation}
         F_{s}^{l+1} = \mathcal{B}^{l}(\mathcal{A}^l(F_{d}^l, F_{s}^l))
\end{equation}
    %  \\
where $\mathcal{A}^l(\cdot)$ is the NCA module and $\mathcal{B}^l(\cdot)$ refers to the CNN block. In the dynamic branch, $F_{d}^{l}$ is processed only by the 2D CNN blocks as it is in the original backbone.
After these two operations, the output $F_{s}^{l}$ is designed to be in the same dimension as $F_{d}^{l}$ for each layer. 
%NCA module is applied to create efficient neighborhood cross attention between intermediate features, while DSI module is used to aggregate the outputs of two branches for the final interaction.
% The input to the first block is the fused features using a NCA directly on BEV features $F_{d}$ and $F_{s}$. 
After all $N$ blocks ($N$ is identical to the number of blocks in the dynamic 2D backbone), the feature maps $F^{out}_{d}$ and $F^{out}_{s}$ have smaller sizes, i.e., the features are dense compared to the BEV features in the beginning. At this stage, the DSI module enhances the interaction between two features, whose output is fed to the rest of the pipeline for object detection. 

\subsection{Dynamic-Static Fusion Module}
%In this section, we introduce the details of the two core components in DynStaF, NCA and DSI. 
%The former one is designed for sparse feature maps to efficiently build the cross attention by only taking neighborhood pixels into consideration, while the latter one performs feature interaction between two branches.
% for the feature fusion in a global manner.

%\paragraph{CNN-based fusion.}
\paragraph{Neighborhood Cross Attention (NCA) module.}
%use query from single frame to find information in multiframe: the fused features are precise in position and rich in semantic information
Considering that BEV features from the static branch provides precise object location information and the rich spatio-temporal semantic information can be found in dynamic branch, we use features in $F_s^l$ as queries and generate keys and values from $F_{d}^l$ to achieve the cross attention.
%Using a vanilla transformer attempts to create the attention between one pixel and all other pixels in a feature map, which establishes the global attention for each pixel. 
% However, BEV feature maps are relative large but sparse, and a few amount of locations are occupied with non-empty pillars. Thus,
As relevant features for the same object should locate at a similar position in both features, we argue that the local information in the neighborhood of a specific query is more essential to build the cross attention in the context of BEV feature maps. Moreover, BEV feature maps are relatively large but sparse, where only a few number of pixels are occupied with non-empty pillars. Limiting a neighborhood helps save computational cost as well.
%Therefore, we argue that the local information in the neighborhood of a specific query is more essential to build the attention in the context of BEV feature maps. 
We adapt Neighborhood Attention Transformer proposed for establishing self attention in the image classification task in \cite{hassani2022neighborhood} to our purpose of building cross attention. 
%NCA only operates on features at these positions, which is a sparse operation specified for point cloud and helps save much computational cost.
The illustration of the cross attention is shown in \Cref{fig:archi} (Bottom Left).

% $A_c \in \mathbb{R}^{D_q\times W \times H}$
Concretely, we first tokenize the BEV feature map using convolutional layers and denote it as a sequence of $m$-dim feature vectors, for instance the feature sequence from the static branch is $F_s \in \mathbb{R}^{n\times m}$. 
%we first tokenize the feature sequences in BEV feature maps using convolutional layers. 
The tokenized feature $F_s$ is linearly projected to a query $Q_s \in \mathbb{R}^{n \times q}$. For the tokenized feature sequence in the multi-frame branch, it is projected to the key $K_d \in \mathbb{R}^{n \times q}$ and the value $V_d \in \mathbb{R}^{n \times v}$ using a linear layer. The cross attention $A_c$ for a query $i$ in the single-frame feature map is calculated as:
\begin{equation}
    A_c^{i} = \sigma(\frac{Q_s^{i}\cdot ({K_d^{\rho(i)}})^T + B^{(i,\rho(i))}}{\sqrt{v}}) \cdot V_d^i
\end{equation}
where $\rho(i)$ is the neighborhood with the size of $k$ centered at the same position in the multi-frame branch, $B^{(i,\rho(i))}$ denotes the positional bias added to the attention and $\sigma$ refers to SoftMax. When multi-headed attention is applied, the outputs of each head are concatenated. For each pixel in the feature map, we calculate the cross attention as above.
%In other word, the cross attention $A_c^{i} \ in \mathbb{R}^{k,q}$ at the position $i$ is the calculated as the dot product of the query from the single-frame with the $k$ nearest neighboring keys at the same position $i$ from the multi-branch. 

\begin{figure}[t]
    \centering
    \includegraphics[width=\linewidth]{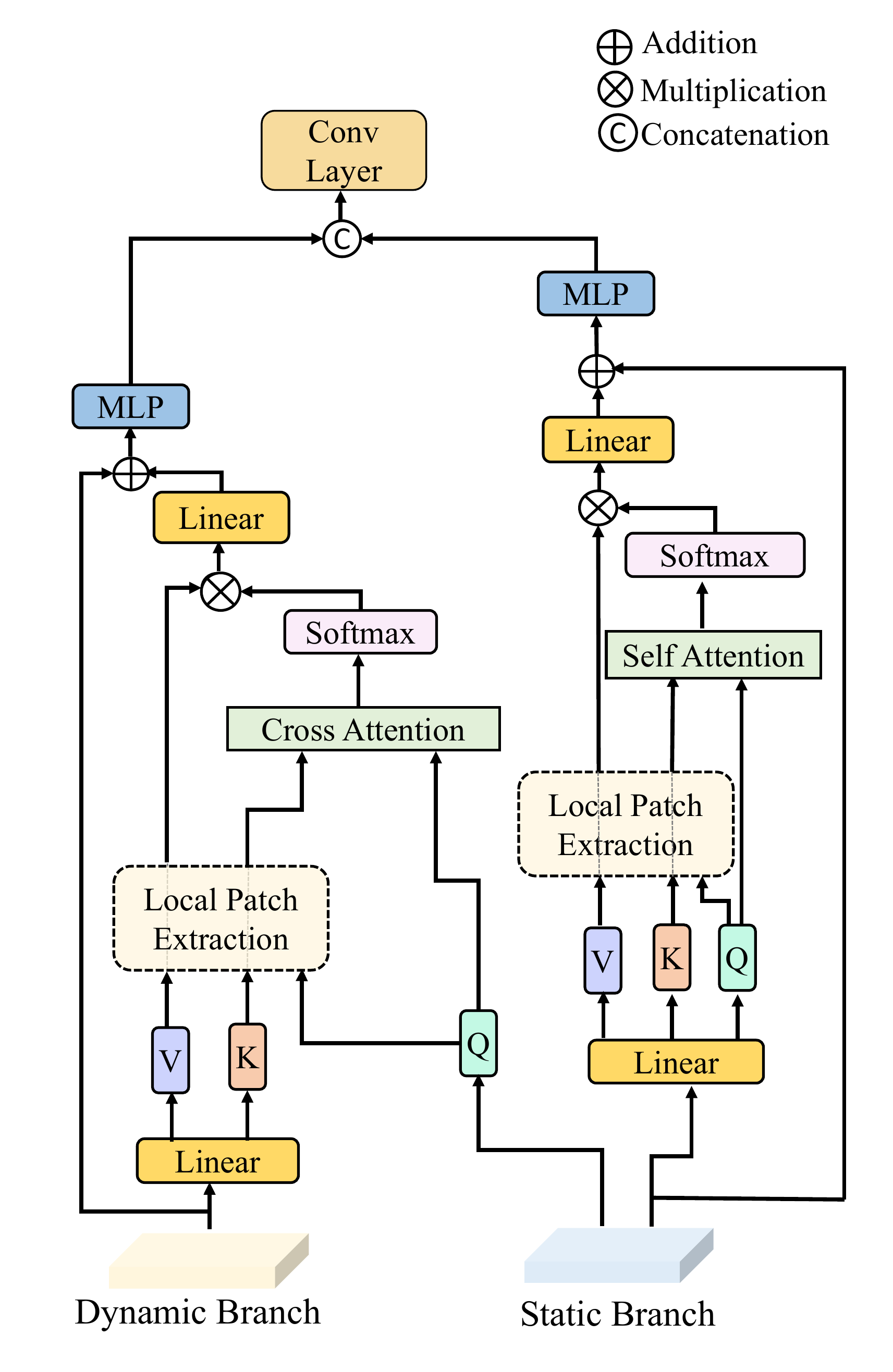}
  \caption{Illustration of Neighborhood Cross Attention (NCA) module. Input is feature maps from two branches $F_s^l$ and $F_d^l$.}
  \label{fig:NCA}
\end{figure}

Another linear layer is added on top of the $A_c^i$. A shortcut and two extra linear layers are utilized to further process this output. To enhance the features with accurate position information, we compute the self attention of the single-frame branch using the same algorithm, but use the linear projection to obtain queries, keys and values all from the single-frame feature. The concatenation of the outputs from the cross attention and self attention is fed into a convolutional layer, leading to the final output of the NCA module. The complete operation of NCA is depicted in \Cref{fig:NCA}. 

% which is a set of indices of pixels nearest to $(i, j)$, the neighborhood cross attention can be represented as follows:

% \textcolor{blue}{CROSS ATTENTION REPRESENTATION}

% where $Q$, $K$ and $V$ are linear projection of corresponding features. We perform this cross attention operation over all grids in single-frame feature map, such that the temporal information is aggregated effectively to the fused feature, which is then precise in position and rich in semantic information. In addition, to make full advantage of point cloud sparsity in BEV, we extend this neighborhood cross attention to sparse one, which means, the number of input tokens can be not fixed, only interested grids are taken into consideration.

\paragraph{Dynamic-Static Interaction (DSI) module.}
After processed by the CNN and NCA blocks, feature maps become dense and rich in information. Although the static branch is already enhanced with local features from dynamic branch by NCA, it cannot guarantee to keep all detailed semantic knowledge of the objects. Therefore, we add an interaction module before the detection head to sufficiently consolidate features from two branches. 
% \textcolor{blue}{To sufficiently consolidate the output features from two branches, an interaction module is introduced before detection head.}
%Note that, for the simplicity, we use the same notation $C, W, H$ to refer to the weight and height dimensions for the feature maps as in the last section, i.e., 
%At this stage, features are fused using convolutional layers. 
Given the features from the single-frame branch denoted as $F_s^{out} \in \mathbb{R}^{C\times W\times H}$ and $F_d^{out} \in \mathbb{R}^{C\times W\times H}$ from the multi-frame branch, the concatenation of both feature maps $F_c\in \mathbb{R}^{2C\times W\times H}$ is used to guide the interaction as it contains a comprehensive view of both features. Specifically, three convolutional layers first process $F_c$, $F_s^{out}$ and $F_d^{out}$ separately, whose output here denote as $F_c'$, $F_s'$ and $F_d'$. DSI takes them as input and produces two feature maps for each branch using CNN blocks as depicted in \Cref{fig:archi} (Bottom Right). The two output components of DSI is then concatenated together with $F_c'$, followed by another CNN block to produce the output $F_o \in \mathbb{R}^{C \times W \times H}$, which is fed into the detection head.

%Then, $F_c'$ is concatenated with $F_s'$ or $F_d'$ to form the merged feature specific to each branch, followed by a simple convolution block containing two CNN layers to obtain the interaction weights, which has the same dimension as the $F_s'$ or $F_d'$.
%$\mathbb{R}^{3C\times W\times H}$. The interaction weights 
% the interaction for the static branch, a simple convolution block containing two CNN layers are used on the concatenation of $F_s'$ and $F_c'$. $A_d\in \mathbb{R}^{3C\times W\times H}$ for the multi-frame branch are calculated likewise. 
%Both attention-weighted feature maps (e.g., $A_d \cdot F_d^{N+1}$) are concatenated with the $F_c'$, The interaction weights is used to guide the feature extraction, which is multiplied with $F_s'$ or $F_c'$ to get features for the interaction. 
%After the concatenation, another CNN block is added on the top to produce the output $F_o \in \mathbb{R}^{C \times W \times H}$, which is fed into the detection head.

\section{Experiments}
\subsection{Implementation details}
\paragraph{Dataset.}
We conduct our experiments on nuScenes \cite{caesar2020nuscenes}, which is a large-scale dataset collected in the real-world containing multiple sensor data. In this work, we only use the LiDAR data (with the frequency of 20 FPS) to tackle with the 3D object detection task. In total, there are 700 video sequences in the training set, 150 videos in validation and test sets each. Following most of the previous works \cite{deng2022vista, caesar2020nuscenes, bai2022transfusion}, we use 10 sweeps for the multi-frame input, corresponding to the LiDAR information in the previous 0.5s. Ten object categories ranging from cars, pedestrians to traffic cones are labelled.

% Some details about model architecture, implementation details of the added single-frame branch on the Pointpillars and CenterPoints.
\paragraph{Architecture details.}
We test our DynStaF strategy on two popular frameworks, Pointpillars \cite{lang2019pointpillars} and CenterPoint \cite{yin2021center}. When adding DynStaF to the PointPillars, we use three NCA modules to get the fused features, the same number of CNN blocks in the 2D backbone as in the original framework. CenterPoint network has two 2D convolutional blocks before the CenterPoint Head, where DynStaF is plugged in. As CenterPoint contains 3D backbone to process voxels, we reduce the channel number in the 2D backbone by half to keep the comparable computation cost. For each NCA module, given the corresponding CNN block in the dynamic branch with the input feature size of $c_{i}\times w_{i}\times h_{i}$ and the output size of $c_{o} \times w_{o} \times h_{o}$, we first use two convolutional layers to tokenize the features (i.e., $F_d^{i}$ and $F_s^{i}$) into the features with the size $\frac{c_{o}}{2} \times w_{i} \times h_{i}$. Then, the cross attention is calculated with the neighborhood range set to 7 and the number of attention heads to 8. Each NCA module produces features with the same output size of $c_{o} \times w_{o} \times h_{o}$.

\paragraph{Training Loss}
We use the anchor-based loss proposed in \cite{lang2019pointpillars} for training the PointPillars-based model. The loss is the weighted sum of three components: localization loss (L1 loss), classification loss (focal loss) and direction loss (cross-entropy loss), whose weights are 0.25, 1.0, 0.2, respectively. The loss used to train the CenterPoint-based model is anchor-free \cite{yin2021center}, which contains classification loss and regression loss. The former one is the cross-entropy loss between the predicted and ground-truth labels with the weight of 1, and the latter one is the L1 regression loss of bounding boxes with the weight of 0.25. %The weight for the classification loss is set to 1 and to 0.25 for the regression loss.

% Training details. 
\paragraph{Training details.}
For PointPillars-based models, the point range is set to [-51.2m, 51.2m] for x-/y-axis and [-5, 3] for z-axis. During training, points are randomly flipped along x- and y-axis. Random rotation with a range of [-$\frac{\pi}{8}$, $\frac{\pi}{8}$] around the z-axis is applied. Moreover, a random global scaling factor is set in the range of [0.95, 1.05]. When uisng CenterPoint with the voxel size of (0.075m, 0.075m, 0.2m), random rotation along z-axis is set to [-$\frac{\pi}{4}$, $\frac{\pi}{4}$]. 
%translated with range [0:2; 0:2; 0:2] m in x,y,z axes.
Class-balanced sampling \cite{zhu2019class} is deployed in all training. 
We follow the same training scheme used in \cite{openpcdet2020,lang2019pointpillars, yin2021center}. Our experiments are conducted under the framework OpenPCDet \cite{openpcdet2020} and all models are trained for 20 epochs with the batch size of 32 on 8 V100 GPUs. 

\paragraph{Evaluation metrics.}
Following the nuScenes benchmark for the detection task \cite{caesar2020nuscenes}, evaluation metrics used in our experiments include mAP (mean Average Precision) and a set of True Positive metrics (TP metrics). When calculating mAP, the criterion for matching between prediction and ground-truth is the 2D center distance on the ground plane. The final mAP score is averaged over all thresholds and all classes. A match in TP metrics is defined as the center distance is inside 2m. There are five TP metrics and the final score for each metric is averaged over all classes (ATE, ASE, AOE, AVE, and AAE measuring translation, scale, orientation, velocity, and attribute errors, respectively). As different metrics capture different performance aspects, a nuScenes detection score (NDS) is defined by combining all these metrics together.

\subsection{Comparison with other methods}
\paragraph{Results on nuScenes validation set.}
We train our model on the nuScenes training set and evaluate on the validation set. \Cref{tab:val} reports the comparison with other state-of-the-art methods. We use mAP and NDS as evaluation metrics. To fairly benchmark different results, we also consider the performance gain of each strategy compared to its own backbone model reported in the original paper, as models may vary in performance with differently trained backbone models. Compared to our re-implemented PointPillars using the multi-frame as input, our DynStaF improves mAP by 5.8\% and NDS by 4.1\%. \cite{bai2022transfusion, deng2022vista, zhou2022centerformer} deploy CenterPoint \cite{yin2021center} as their backbone model. We see that our DynStaF achieves the most performance gain in both metrics compared to all other SOTA methods using the CenterPoint as the backbone. Our CP+DynStaF reaches 67.1\% and 58.9\% on NDS and mAP, respectively, which achieves the best performance on the validation set. Performance gain is significant on both backbone models, highlighting the compatibility of DynStaF.

\paragraph{Results on nuScenes test set.}
Besides the offline evaluation, we compare DynStaF with other SOTA single models on the nuScenes test server. No Test Time Augmentation (TTS) was used during the test phase. For a fair comparison, we compare with the results without TTS in \Cref{tab:test}. The methods are divided into two groups: (1) pillar-based methods which do not contain 3D convolutional operations; (2) voxel-based methods with 3D convolution blocks. As previous pillar-based methods usually deploy single-frame only as input, we also include a baseline for our re-implemented PointPillars with multi-frame as the input for a fair comparison. When using multi-frames, the vanilla PointPillar achieves 44.6\% in mAP and 57.7\% in NDS. With our DynStaF, PointPillars is improved by a large margin (5.9\% in mAP and 3.9\% in NDS), leading to the state-of-the-art performance for the pillar-based model: 50.5\% for mAP and 61.6\% for NDS.
Moreover, notable improvement on individual object category can be observed. For example, on the traffic cone and barrier, DynStaF increases the mAP compared to the previous best results by 6.3\% and 12.5\%, respectively. Our DynStaF significantly strengthens pillar-based backbone, narrowing the performance gap compared to the voxel-based methods. 

When comparing with other methods using 3D convolutional blocks, our DynStaF achieves the best performance in both mAP with 61.0\% and NDS with 67.7\%. When compared to the backbone method CenterPoint \cite{yin2021center}, DynStaF increases its performance in mAP by 3.0\% and in NDS by 2.2\%, which indicates its effectiveness. In particular, mAP of the construction vehicle or motorcycle is improved by a large margin. Overall, CenterPoint+DynStaF achieves the state-of-the-art performance on the nuScenes test set without bells and whistles. 

\begin{table}[t]
   \centering
   \resizebox{\linewidth}{!}{
   \begin{tabular}{c|cccc}
   \toprule
    & \textbf{mAP} & \begin{tabular}[c]{@{}c@{}} \textbf{mAP} \\ Gain \end{tabular} & \textbf{NDS} & \begin{tabular}[c]{@{}c@{}} \textbf{NDS} \\ Gain \end{tabular}\\ 
   \toprule
   PP* \cite{lang2019pointpillars} (CVPR 19) & 43.7 & - & 57.3 & - \\
   PP + DynStaF (ours) & 49.4 &  \textbf{5.8} & 61.4 & \textbf{4.1} \\\midrule \midrule
   %DETR + DCA & 47.01 & 60.04\\
   CP* \cite{yin2021center} (CVPR 21) & 58.0 & - & 65.7 & - \\
  %TransFusion-L \cite{bai2022transfusion} &  \textbf{60.0} &  2.6 & 66.8  &  1.6 \\
   CP + VISTA \cite{deng2022vista} (CVPR 22) &57.6 & 1.2 & 65.6 & 0.8 \\
  CenterFormer \cite{zhou2022centerformer} (ECCV 22) & 55.4 & 0.2  & 65.2  & 0.8 \\
  %MDRNet \cite{huang2022rethinking} (arXiv 22) & 61.2 & 1.6 &  67.9 & 1.2\\
  CP + DynStaF (ours) &  \textbf{58.9} & 0.9 &  \textbf{67.1} & 2.2 \\
  
  \bottomrule
   \end{tabular}
   }
    \caption{Comparison with other SOTA methods on nuScenes validation set. * denotes our re-implementation backbone results. Result of mAP/NDS and its performance gain (compared to implemented backbones reported in previous works) are listed.} 
   \label{tab:test}
\end{table}

%%%%%%%%%%%%%%%%%%%
\begin{table*}[]
   \centering
   \resizebox{\linewidth}{!}{
   \begin{tabular}{c|cc|cccccccccc}
   \toprule
    %& reference 
    & \textbf{mAP} & \textbf{NDS}  & car & truck & bus &  trailer & cons. & pedest. & motor. & bicycle & traff. & barrier  \\ 
   \toprule
   PointPillars \cite{lang2019pointpillars} %& CVPR 19 
   & 30.9 & 45.3 & 68.4 & 23.0 & 28.2 & 23.4 & 4.1 & 59.7 & 27.4 & 1.1 & 30.8 & 38.9 \\
   WYSIWYG \cite{hu2020you} %& CVPR 20
   & 41.9 & 35.0 & 79.1 & 30.4 & 46.6 & 40.1 & 7.1 & 65.0 & 18.2 & 0.1 & 28.8 & 34.7 \\ 
   InfoFocus \cite{wang2020infofocus} %& ECCV 20 
   & 39.5 & - & 77.9 & 31.4 & 44.8 & 37.3 & 10.7 & 63.4 & 29.0 & 6.1 & 46.5& 47.8 \\
   PMPNet \cite{yin2020lidar} %& CVPR 20 
   & - & 45.4 & 79.7& 33.6 & 47.1 & 43.0 & 18.1 & 76.5 & 40.7 & 12.3 & 58.8 & 48.4 \\
   PointPillars \cite{lang2019pointpillars}(Multi) * %&  
   & 44.6& 57.7& 80.3 & 44.7 & 55.5  & 47.6  & 11.4 & 69.7  & 27.7 & 5.4 & 54.4 & 49.7  \\
   PP + DynStaF (ours)
   & \textbf{50.5} & \textbf{61.6} & 82.3 & 46.3 & 56.0 & 51.9 & 14.2 & 74.9 & 41.2 & 10.8 & 65.1 & 62.2 \\\midrule \midrule
%   CenterPoint \cite{yin2021center} & CVPR 21 & 60.3 & 67.3 & 85.2 &  53.5 & 63.6 & 56.0 & 20.0 & 84.6 & 59.5 & 30.7 & 78.4 & 71.1  \\
   CGBS \cite{zhu2019class} %& arXiv 19 
   & 52.8 & 63.6 & 81.1 & 48.5 & 54.9 & 42.9 & 10.5 & 80.1 & 22.3 & 22.3 & 70.9 & 65.7 \\
   Pointformer \cite{pan20213d}& 53.6 & - & 82.3 & 48.1 & 55.6 & 43.4 & 8.6 & 81.8 & 55.0 & 22.7& 72.2 & 66.0\\
  %TransFusion-L \cite{bai2022transfusion} &  \textbf{60.0} &  2.6 & 66.8  &  1.6 \\
  %CenterFormer \cite{zhou2022centerformer} (ECCV 22) & 55.4 & 0.2  & 65.2  & 0.8 \\
  %MDRNet \cite{huang2022rethinking} (arXiv 22) & 61.2 & 1.6 &  67.9 & 1.2\\
  CVCNet \cite{chen2020every} & 55.8 & 64.2 & 82.6 & 49.5 & 59.4 & 51.1 & 16.2 & 83.0 & 61.8 & 38.8 & 69.7 & 69.7 \\
  CenterPoint \cite{yin2021center} %& CVPR 21
  &  58.0 & 65.5 & 84.6 & 51.0 & 60.2 & 53.2 & 17.5 & 83.4 & 53.7 & 28.7 & 76.7 & 70.9 \\
  OHS \cite{chen2020object} %& ECCV 20 
  & 59.3 & 66.0 & 83.1 & 50.9 & 56.4 & 53.3 & 23.0 & 81.3 & 63.5 & 36.6 & 73.0 & 71.6 \\
%   CenterPoint $\dagger$ \cite{yin2021center} %& CVPR 21 
%   & 60.3 & 67.3 & 85.2 &  53.5 & 63.6 & 56.0 & 20.0 & 84.6 & 59.5 & 30.7 & 78.4 & 71.1  \\
%   OHS + VISTA $\dagger$ \cite{deng2022vista} & CVPR 22 & 63.0 & 69.8 & 84.4 & 55.1 & 63.7 & 54.2 & 25.1 & 82.8 & 70.0 & 45.4 & 78.5 & 71.4 \\
  CP + DynStaF (ours) & \textbf{61.0} & \textbf{67.7} & 84.6 & 51.2 & 61.2 & 56.5 & 23.5 & 85.0 & 64.7 & 32.3 & 80.3 & 70.5   \\
  
  \bottomrule
   \end{tabular}
   }
    \caption{Comparison with other SOTA (non-ensemble) LiDAR-based methods on nuScenes test set (without Test Time Augmentation). ``cons.'',  ``pedest.'', ``motor.'' and ``traff.'' refer to construction vehicle, pedestrian, motorcycle and traffic cone, respectively. The first block is the methods not utilizing 3D convolutional networks, while the second block is.  * denotes our re-implementation results.} 
   \label{tab:val}
\end{table*}
%%%%%%%%%%%%%%%%%%%%

\subsection{Ablation study}
In this section, we thoroughly analyze the effectiveness of each component, i.e., NCA, DSI and dual pathway architecture in our fusion strategy. All ablation studies are conducted on the NuScenes validation set and using the PointPillars \cite{lang2019pointpillars} as the baseline model. The results are listed in \Cref{tab:ablation}. Using multi-frame point cloud as input, PointPillars without any feature fusion achieves 57.33\% for NDS and 43.66\% for mAP, respectively. If we use the naive feature fusion (denoted as ``CNN-only'') to replace the NCA module after each block, i.e., concatenating two features and adding CNN layers on top of it, the performance is improved to 59.74\% NDS, which verifies that both feature branches have complementary information. With a more sophisticated fusion module, our proposed NCA module, the NDS is improved further to 60.53\% and mAP is increased by a large margin (4.27\%) compared to the baseline. This indicates that using transformer-based cross attention mechanism is effective in the context of sparse point clouds. When the feature maps are dense, using DSI is more effective, as we see that NCA-only fusion is inferior to our final approach NCA + DSI. %Compared to using NCA for the last layer of feature fusion, using DSI improves the performance by 1.49\% and 0.88\% on mAP and NDS, respectively.

%%%%%%%%%%%%%%%%%%%%%
\begin{table}[b]
   \centering
   \begin{tabular}{c|cc}
   \toprule
    & \textbf{mAP} & \textbf{NDS} \\ 
   \midrule
   Pointpillar* \cite{lang2019pointpillars} & 43.66  & 57.33 \\\midrule
   CNN-only fusion & 47.49 & 59.74  \\
   NCA-only fusion & 47.93 & 60.53 \\\midrule
   %DETR + DCA & 47.01 & 60.04\\
   NCA + DSI (Single) & 2.70 & 19.35  \\
   NCA + DSI (Dual) & \textbf{49.42} & \textbf{61.41}   \\\bottomrule
   \end{tabular}
    \caption{Ablation study results on nuScenes validation set. * denotes our re-implementation.} 
   \label{tab:ablation}
\end{table}
%%%%%%%%%%%%%%%%%%%%%%%%
We study the effectiveness of the dual-pathway architecture in the second block in \Cref{tab:ablation}. Instead of two feature streams, only one single pathway for feature fusion is deployed, i.e., the single-frame and multi-frame branch share weights of 2D CNN blocks highlighted by the color gray in \Cref{fig:archi}. The poor performance of this model (denoted as ``Single") proves that it is impossible for a single backbone to deal with the single-frame and multi-frame features simultaneously. This also reveals that the multi-frame and single-frame contain different information. %Therefore, feature fusion can enhance the performance by combining the information reasonably. 
Our DynStaF (``Dual'') arrives the best performance at 49.42\% on mAP and 61.41\% on NDS using all components, demonstrating the advantage of our proposed feature fusion strategy.

\subsection{Analysis}
\paragraph{Other cross-attention modules.}
Deformable DETR proposed in \cite{zhu2020deformable} learns to attend to a small set of keys around a reference point, which similarly discovers local attention as our NCA. The difference is that the sampling offsets (the position of keys) is learnable in Deformable attention module, while our NCA produces the ``global'' attention within a neighborhood. Deformable DETR has been proven to be efficient in the feature fusion based on LiDAR point clouds such as in \cite{li2022bevformer,zhou2022centerformer}.
%for instance BEVformer \cite{li2022bevformer} uses deformable DETR to learn the cross attention between different camera views. Moreover, it is applied to extract temporal attention between the current image frame and the previous one. Similarly, Deformable DETR in CenterFormer \cite{zhou2022centerformer} extracts the multi-scale cross attention between the current BEV feature map with the previous ones. 
To discover the capability of the deformable attention in our use case, we replaced the NCA module in the pillar-based DynStaF with the Deformable DETR layer. However, we saw the performance degradation: NDS decreased to 60.04\% and the mAP dropped to 47.01\%. It indicates that the whole neighborhood is essential to build the cross attention between two branches. 

Moreover, we also explored some other methods for the final interaction of features from two branches. For instance, we adapted the CBAM module \cite{woo2018cbam} to learn the cross attention between two branches. When replacing DSI in the pillar-based DynStaF, NDS and mAP declined to 60.70\% and 48.42\%, respectively. These results show the advantages of our NCA and DSI in aggregating features and enhancing the interaction between two branches.
\paragraph{Efficiency of static branch.} In the CenterPoint-based DynStaF, we aggregate the features in the 2D convolutional blocks and keep the channel dimensions in the two blocks the half as in the CenterPoint. We found this setting was not only efficient but also advantageous in the final performance. As we used an identical branch as the dynamic branch, i.e., a 3D convolutional backbone was used for single-frame input and the channel dimension was set to the same as in the original CenterPoint. We got 66.28\% NDS and 58.41\% mAP, which were inferior to the results of our DynStaF. The reason could be that the single-frame input was not sufficient to train a complex backbone. The current design of DynStaF provides lower computational cost and satisfactory performance in 3D detection at the same time.
%As shown in \Cref{fig:archi}, the feature fusion happens at different stages in the 2D BEV backbone. To show the effectiveness of this design, we remove the first fusion block before sending to the original BEV feature encoder.
%\setlength{\abovecaptionskip}{2mm}
%\setlength{\belowcaptionskip}{-6mm}
\begin{figure}[t]
    \centering
  \begin{subfigure}[b]{0.45\linewidth}
    \includegraphics[width=\textwidth]{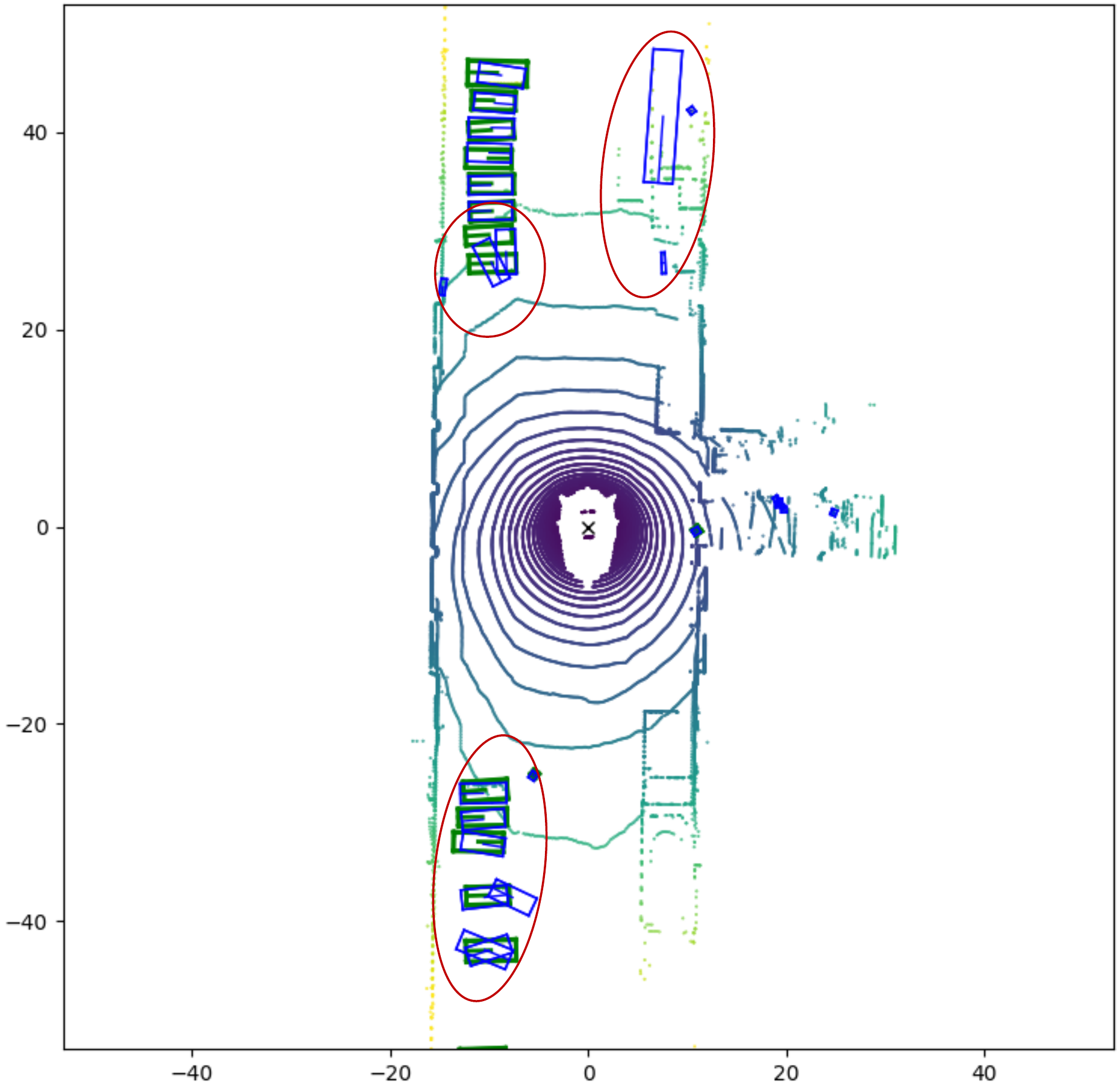}
    \caption{}
    \label{fig:cp1}
  \end{subfigure}
  \hfill
  \begin{subfigure}[b]{0.45\linewidth}
    \includegraphics[width=\textwidth]{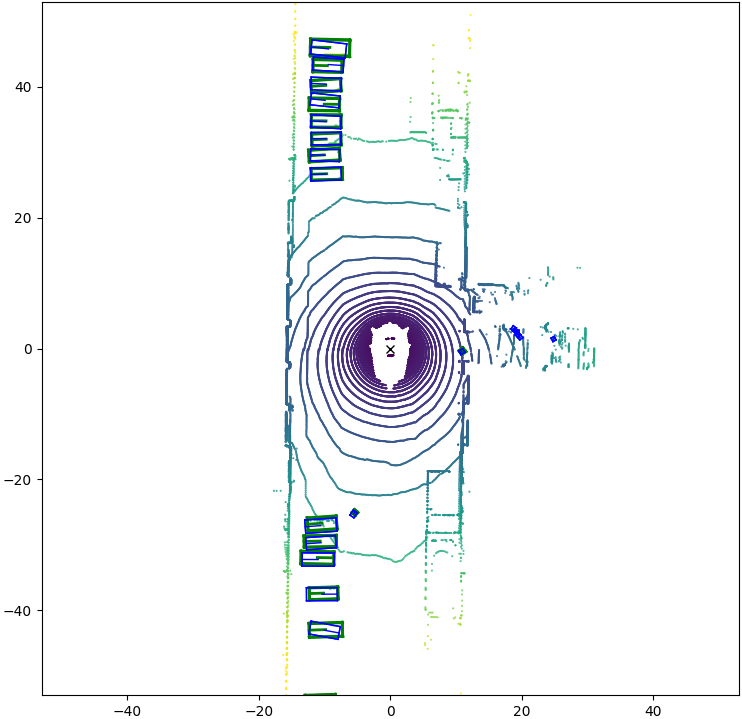}
    \caption{}
    \label{fig:dynstaf1}
  \end{subfigure}
\\
    \begin{subfigure}[b]{0.45\linewidth}
    \includegraphics[width=\textwidth]{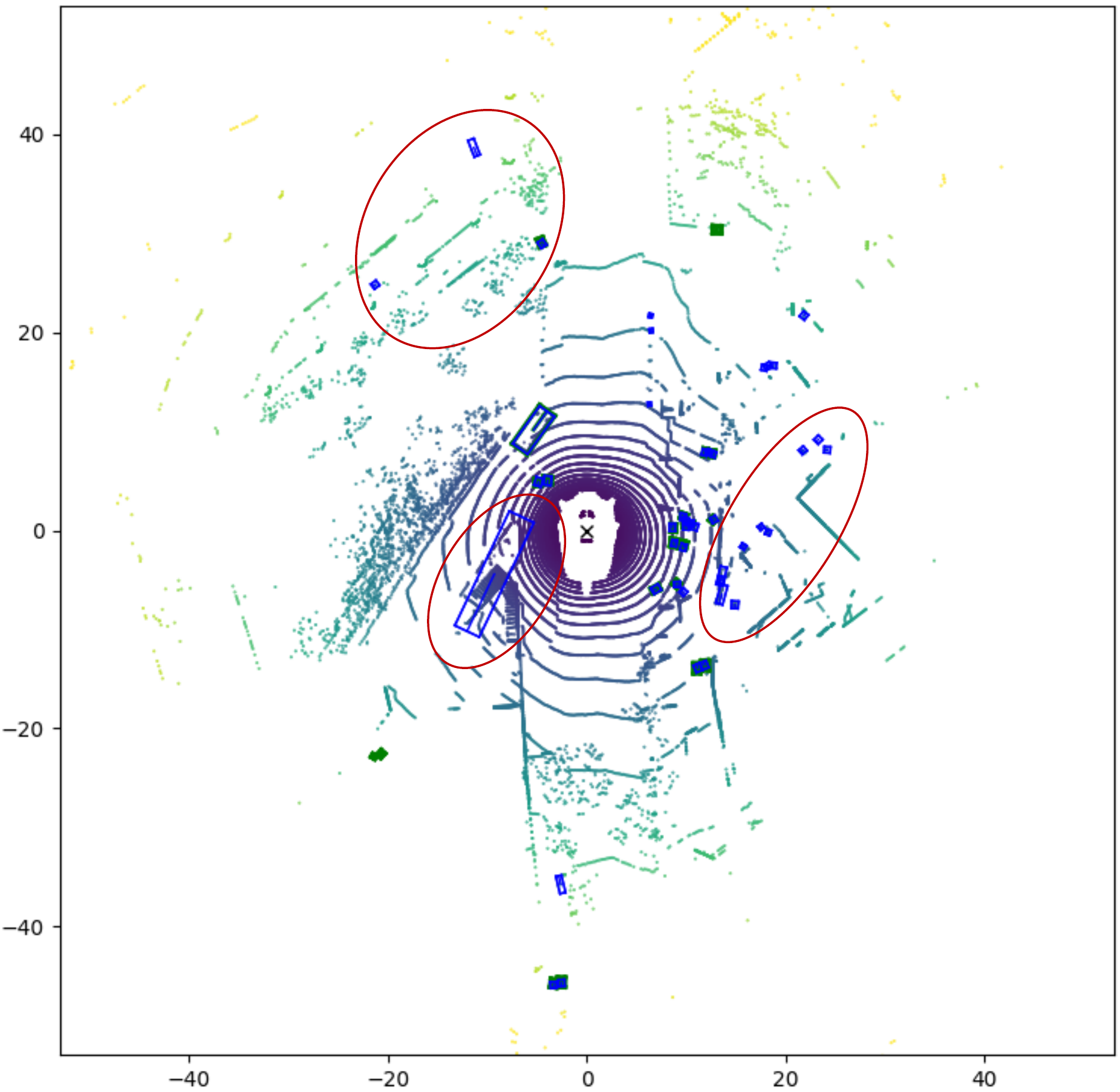}
    \caption{}
    \label{fig:cp2}
  \end{subfigure}
  \hfill
  \begin{subfigure}[b]{0.45\linewidth}
    \includegraphics[width=\textwidth]{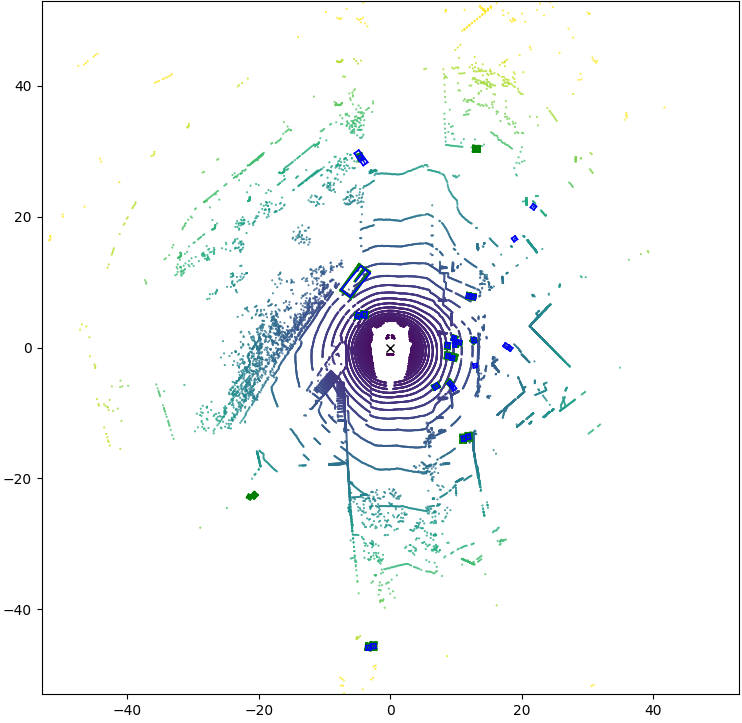}
    \caption{}
    \label{fig:dynstaf2}
  \end{subfigure}
  \caption{Visualization of object detection results on nuScenes validation set. Each row refers to one sample. (a) and (c): CenterPoint w/o DynStaF; (b) and (d): CenterPoint with DynStaF. Ground-truth bounding boxes are in green and prediction bounding boxes in blue.}
  \label{fig:viz pred}
\end{figure}

\subsection{Qualitative Results}
We qualitatively show the advantage of DynStaF in the 3D object detection task. \Cref{fig:viz pred} shows the prediction using CenterPoint as the backbone. In the first scene where there is a queue of vehicles on the left side. CenterPoint cannot detect the position of several vehicles (marked in the red circles) precisely, as shown in \Cref{fig:cp1}. DynStaF in \Cref{fig:dynstaf1} localizes these vehicles correctly. Furthermore, DynStaF alleviates the false positives compared to the baseline. The second example is collected on a city street surrounded by many buildings with a group of pedestrian walking on the back right side of the car. As discussed in \Cref{fig:cp1}, multi-frame input is overwhelmed with point clouds in this case, making the detection difficult. For instance in \Cref{fig:cp2}, the point cloud of the walking pedestrians will be crowded such that the model predicts falsely (marked with the red circle on the right). With DynStaF, the single-frame can provide a clearer view of the each pedestrian as the point clouds are more sparse. These two examples show the advantage of our DynStaF in predicting location precisely and avoiding false positive detection.

%where no vehicles on the right side in the ground-truth, without DynStaF there are many objects falsely detected on the right as shown in \Cref{fig:cp1}. With DynStaF in \Cref{fig:dynstaf1}, the model does not suffer from these false positive predictions. As discussed in \Cref{fig:teaser figure}, the crowded points cloud leads to false positive detection of objects. 
% These two examples validates the advantage of DynStaF in precisely detecting objects by mitigating 
% In the second scenario, the car is driving through a city street with a lot of vehicles are parking on the left side. We notice that with DynStaF
% However, when confronting a complex scene where occlusions are happening with moving objects.  

\begin{figure}[t]
    \centering
    \includegraphics[width=\linewidth]{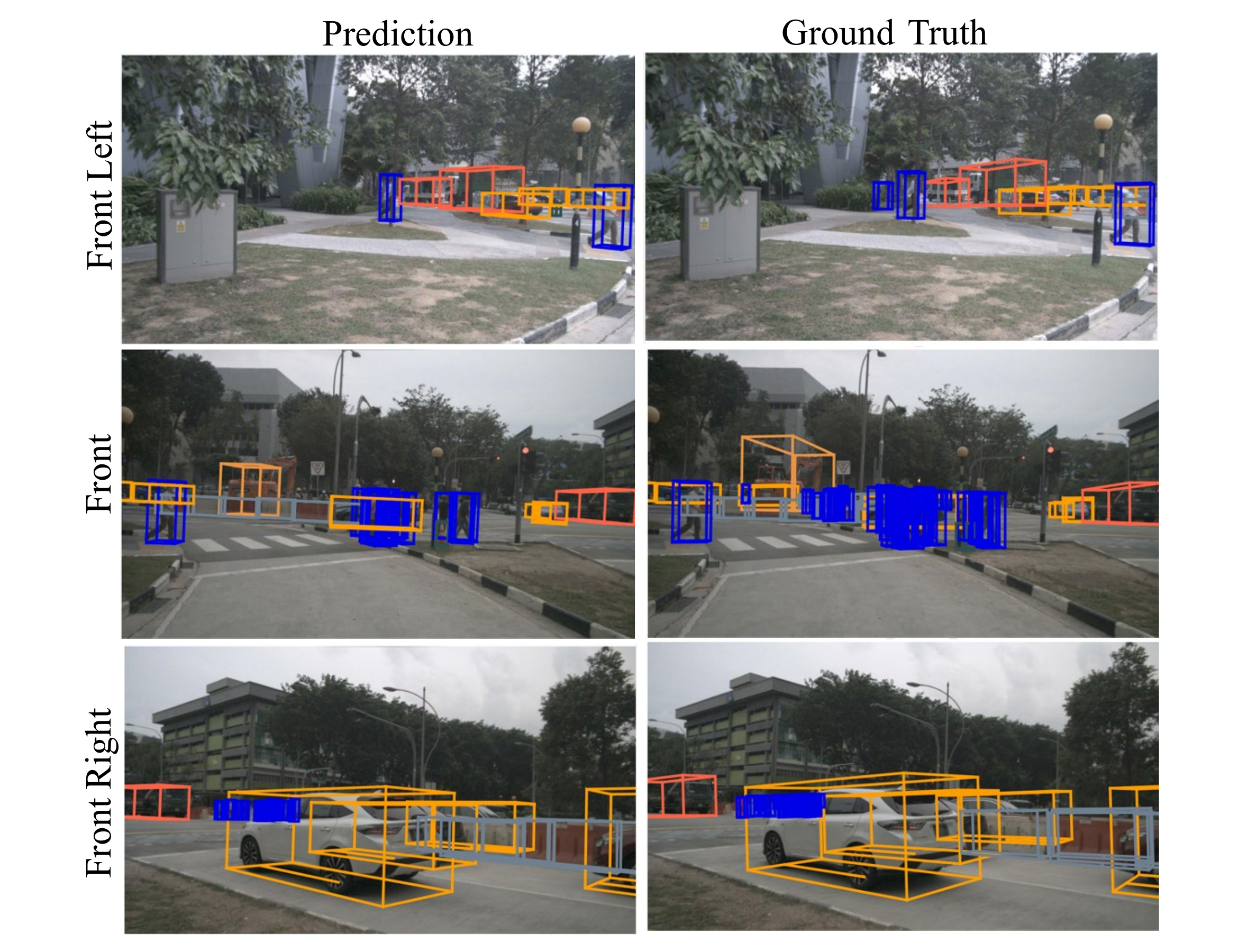}
  \caption{Visualization of object detection results on nuScenes validation set. Different rows represent different camera positions. The first column represents the prediction using CenterPoint+DynStaF; The second column is the ground-truth.}
  \label{fig:cam pred}
\end{figure}

\Cref{fig:cam pred} demonstrates the prediction in a camera view, which highlights concretely the ability of DynStaF in the context of dense point clouds. We show three challenging views where objects are closed to each other. In the front left and front right view, occlusion of vehicles can be observed, and our prediction is correct for most of the objects. In the front view, in which there exists more occlusion such as a group of walking pedestrians, DynStaF detects all objects but it cannot handle the occlusion position perfectly.
 %our model performs well in the scenes containing less occlusion such as in the front left and front right view.

%\setlength{\abovecaptionskip}{2mm}
%\setlength{\belowcaptionskip}{-6mm}

\section{Conclusion}
%As multi-frame LiDAR input becomes popular in 3D object detection because of its rich semantic and motion information, but it loses 
In this work, we propose a novel feature fusion framework named DynStaF to fuse the multi-frame and single-frame LiDAR point clouds for 3D object detection. Neighborhood Cross Attention module in DynStaF fuses features with a limited neighborhood instead of considering global attention, followed by Dynamic-Static Interaction module enhancing the feature interaction. Without loss of generality, our method can be utilized as a plug-and-play module in different pillar-based or voxel-based LiDAR point cloud detection algorithms. Quantitative results show that our DynStaF improves the strong backbone CenterPoint, outperforming other methods on the nuScenes dataset. Qualitatively, we demonstrate that our DynStaF can precisely localize objects thus avoid false positive predictions. Moreover, DynStaF enhanced the real-time pillar-based backbone significantly in the performance, highlighting its potential in practical usages. For future work, we aim to combine DynStaF with powerful frameworks taking LiDAR and camera images as input to further improve 3D object detection.
{\small

\bibliographystyle{plainnat}
\bibliography{egbib}

\begin{thebibliography}{30}
\providecommand{\natexlab}[1]{#1}
\providecommand{\url}[1]{\texttt{#1}}
\expandafter\ifx\csname urlstyle\endcsname\relax
  \providecommand{\doi}[1]{doi: #1}\else
  \providecommand{\doi}{doi: \begingroup \urlstyle{rm}\Url}\fi

\bibitem[Bai et~al.(2022)Bai, Hu, Zhu, Huang, Chen, Fu, and
  Tai]{bai2022transfusion}
Xuyang Bai, Zeyu Hu, Xinge Zhu, Qingqiu Huang, Yilun Chen, Hongbo Fu, and
  Chiew-Lan Tai.
\newblock Transfusion: Robust lidar-camera fusion for 3d object detection with
  transformers.
\newblock In \emph{CVPR}, 2022.

\bibitem[Caesar et~al.(2020)Caesar, Bankiti, Lang, Vora, Liong, Xu, Krishnan,
  Pan, Baldan, and Beijbom]{caesar2020nuscenes}
Holger Caesar, Varun Bankiti, Alex~H Lang, Sourabh Vora, Venice~Erin Liong,
  Qiang Xu, Anush Krishnan, Yu~Pan, Giancarlo Baldan, and Oscar Beijbom.
\newblock nuscenes: A multimodal dataset for autonomous driving.
\newblock In \emph{CVPR}, 2020.

\bibitem[Carion et~al.(2020)Carion, Massa, Synnaeve, Usunier, Kirillov, and
  Zagoruyko]{carion2020end}
Nicolas Carion, Francisco Massa, Gabriel Synnaeve, Nicolas Usunier, Alexander
  Kirillov, and Sergey Zagoruyko.
\newblock End-to-end object detection with transformers.
\newblock In \emph{ECCV}, 2020.

\bibitem[Chen et~al.(2020{\natexlab{a}})Chen, Sun, Cheung, and
  Yuille]{chen2020every}
Qi~Chen, Lin Sun, Ernest Cheung, and Alan~L Yuille.
\newblock Every view counts: Cross-view consistency in 3d object detection with
  hybrid-cylindrical-spherical voxelization.
\newblock 2020{\natexlab{a}}.

\bibitem[Chen et~al.(2020{\natexlab{b}})Chen, Sun, Wang, Jia, and
  Yuille]{chen2020object}
Qi~Chen, Lin Sun, Zhixin Wang, Kui Jia, and Alan Yuille.
\newblock Object as hotspots: An anchor-free 3d object detection approach via
  firing of hotspots.
\newblock In \emph{ECCV}, 2020{\natexlab{b}}.

\bibitem[Chen et~al.(2021)Chen, Vora, and Beijbom]{chen2021polarstream}
Qi~Chen, Sourabh Vora, and Oscar Beijbom.
\newblock Polarstream: Streaming object detection and segmentation with polar
  pillars.
\newblock 2021.

\bibitem[Deng et~al.(2022)Deng, Liang, Sun, and Jia]{deng2022vista}
Shengheng Deng, Zhihao Liang, Lin Sun, and Kui Jia.
\newblock Vista: Boosting 3d object detection via dual cross-view spatial
  attention.
\newblock In \emph{CVPR}, 2022.

\bibitem[Hassani et~al.(2022)Hassani, Walton, Li, Li, and
  Shi]{hassani2022neighborhood}
Ali Hassani, Steven Walton, Jiachen Li, Shen Li, and Humphrey Shi.
\newblock Neighborhood attention transformer.
\newblock \emph{arXiv preprint arXiv:2204.07143}, 2022.

\bibitem[Hu et~al.(2020)Hu, Ziglar, Held, and Ramanan]{hu2020you}
Peiyun Hu, Jason Ziglar, David Held, and Deva Ramanan.
\newblock What you see is what you get: Exploiting visibility for 3d object
  detection.
\newblock In \emph{CVPR}, 2020.

\bibitem[Huang et~al.(2022)Huang, Chen, Ding, Liao, Liu, Wu, Nie, Liu, and
  Wang]{huang2022rethinking}
Dihe Huang, Ying Chen, Yikang Ding, Jinli Liao, Jianlin Liu, Kai Wu, Qiang Nie,
  Yong Liu, and Chengjie Wang.
\newblock Rethinking dimensionality reduction in grid-based 3d object
  detection.
\newblock \emph{arXiv preprint arXiv:2209.09464}, 2022.

\bibitem[Lang et~al.(2019)Lang, Vora, Caesar, Zhou, Yang, and
  Beijbom]{lang2019pointpillars}
Alex~H Lang, Sourabh Vora, Holger Caesar, Lubing Zhou, Jiong Yang, and Oscar
  Beijbom.
\newblock Pointpillars: Fast encoders for object detection from point clouds.
\newblock In \emph{CVPR}, 2019.

\bibitem[Li et~al.(2022{\natexlab{a}})Li, Chen, Qi, Li, Sun, and
  Jia]{li2022unifying}
Yanwei Li, Yilun Chen, Xiaojuan Qi, Zeming Li, Jian Sun, and Jiaya Jia.
\newblock Unifying voxel-based representation with transformer for 3d object
  detection.
\newblock 2022{\natexlab{a}}.

\bibitem[Li et~al.(2022{\natexlab{b}})Li, Wang, Li, Xie, Sima, Lu, Yu, and
  Dai]{li2022bevformer}
Zhiqi Li, Wenhai Wang, Hongyang Li, Enze Xie, Chonghao Sima, Tong Lu, Qiao Yu,
  and Jifeng Dai.
\newblock Bevformer: Learning bird's-eye-view representation from multi-camera
  images via spatiotemporal transformers.
\newblock \emph{arXiv preprint arXiv:2203.17270}, 2022{\natexlab{b}}.

\bibitem[Liu et~al.(2022)Liu, Tang, Amini, Yang, Mao, Rus, and
  Han]{liu2022bevfusion}
Zhijian Liu, Haotian Tang, Alexander Amini, Xinyu Yang, Huizi Mao, Daniela Rus,
  and Song Han.
\newblock Bevfusion: Multi-task multi-sensor fusion with unified bird's-eye
  view representation.
\newblock \emph{arXiv preprint arXiv:2205.13542}, 2022.

\bibitem[Noh et~al.(2021)Noh, Lee, and Ham]{noh2021hvpr}
Jongyoun Noh, Sanghoon Lee, and Bumsub Ham.
\newblock Hvpr: Hybrid voxel-point representation for single-stage 3d object
  detection.
\newblock In \emph{CVPR}, 2021.

\bibitem[Pan et~al.(2021)Pan, Xia, Song, Li, and Huang]{pan20213d}
Xuran Pan, Zhuofan Xia, Shiji Song, Li~Erran Li, and Gao Huang.
\newblock 3d object detection with pointformer.
\newblock In \emph{CVPR}, 2021.

\bibitem[Qi et~al.(2017)Qi, Yi, Su, and Guibas]{qi2017pointnet++}
Charles~Ruizhongtai Qi, Li~Yi, Hao Su, and Leonidas~J Guibas.
\newblock Pointnet++: Deep hierarchical feature learning on point sets in a
  metric space.
\newblock 2017.

\bibitem[Sun et~al.(2020)Sun, Kretzschmar, Dotiwalla, Chouard, Patnaik, Tsui,
  Guo, Zhou, Chai, Caine, Vasudevan, Han, Ngiam, Zhao, Timofeev, Ettinger,
  Krivokon, Gao, Joshi, Zhang, Shlens, Chen, and Anguelov]{Sun_2020_CVPR}
Pei Sun, Henrik Kretzschmar, Xerxes Dotiwalla, Aurelien Chouard, Vijaysai
  Patnaik, Paul Tsui, James Guo, Yin Zhou, Yuning Chai, Benjamin Caine, Vijay
  Vasudevan, Wei Han, Jiquan Ngiam, Hang Zhao, Aleksei Timofeev, Scott
  Ettinger, Maxim Krivokon, Amy Gao, Aditya Joshi, Yu~Zhang, Jonathon Shlens,
  Zhifeng Chen, and Dragomir Anguelov.
\newblock Scalability in perception for autonomous driving: Waymo open dataset.
\newblock In \emph{CVPR}, 2020.

\bibitem[Team(2020)]{openpcdet2020}
OpenPCDet~Development Team.
\newblock Openpcdet: An open-source toolbox for 3d object detection from point
  clouds.
\newblock \url{https://github.com/open-mmlab/OpenPCDet}, 2020.

\bibitem[Wang et~al.(2020{\natexlab{a}})Wang, Lan, Gao, and
  Davis]{wang2020infofocus}
Jun Wang, Shiyi Lan, Mingfei Gao, and Larry~S Davis.
\newblock Infofocus: 3d object detection for autonomous driving with dynamic
  information modeling.
\newblock In \emph{ECCV}, 2020{\natexlab{a}}.

\bibitem[Wang et~al.(2020{\natexlab{b}})Wang, Fathi, Kundu, Ross, Pantofaru,
  Funkhouser, and Solomon]{wang2020pillar}
Yue Wang, Alireza Fathi, Abhijit Kundu, David~A Ross, Caroline Pantofaru, Tom
  Funkhouser, and Justin Solomon.
\newblock Pillar-based object detection for autonomous driving.
\newblock In \emph{ECCV}, 2020{\natexlab{b}}.

\bibitem[Woo et~al.(2018)Woo, Park, Lee, and Kweon]{woo2018cbam}
Sanghyun Woo, Jongchan Park, Joon-Young Lee, and In~So Kweon.
\newblock Cbam: Convolutional block attention module.
\newblock In \emph{ECCV}, 2018.

\bibitem[Yan et~al.(2018)Yan, Mao, and Li]{yan2018second}
Yan Yan, Yuxing Mao, and Bo~Li.
\newblock Second: Sparsely embedded convolutional detection.
\newblock \emph{Sensors}, 2018.

\bibitem[Ye et~al.(2022)Ye, Zhou, Chen, Xie, Wang, Wang, and
  Foroosh]{ye2022lidarmultinet}
Dongqiangzi Ye, Zixiang Zhou, Weijia Chen, Yufei Xie, Yu~Wang, Panqu Wang, and
  Hassan Foroosh.
\newblock Lidarmultinet: Towards a unified multi-task network for lidar
  perception.
\newblock \emph{arXiv preprint arXiv:2209.09385}, 2022.

\bibitem[Yin et~al.(2020)Yin, Shen, Guan, Zhou, and Yang]{yin2020lidar}
Junbo Yin, Jianbing Shen, Chenye Guan, Dingfu Zhou, and Ruigang Yang.
\newblock Lidar-based online 3d video object detection with graph-based message
  passing and spatiotemporal transformer attention.
\newblock In \emph{CVPR}, 2020.

\bibitem[Yin et~al.(2021)Yin, Zhou, and Krahenbuhl]{yin2021center}
Tianwei Yin, Xingyi Zhou, and Philipp Krahenbuhl.
\newblock Center-based 3d object detection and tracking.
\newblock In \emph{CVPR}, 2021.

\bibitem[Zhou and Tuzel(2018)]{zhou2018voxelnet}
Yin Zhou and Oncel Tuzel.
\newblock Voxelnet: End-to-end learning for point cloud based 3d object
  detection.
\newblock In \emph{CVPR}, 2018.

\bibitem[Zhou et~al.(2022)Zhou, Zhao, Wang, Wang, and
  Foroosh]{zhou2022centerformer}
Zixiang Zhou, Xiangchen Zhao, Yu~Wang, Panqu Wang, and Hassan Foroosh.
\newblock Centerformer: Center-based transformer for 3d object detection.
\newblock In \emph{ECCV}, 2022.

\bibitem[Zhu et~al.(2019)Zhu, Jiang, Zhou, Li, and Yu]{zhu2019class}
Benjin Zhu, Zhengkai Jiang, Xiangxin Zhou, Zeming Li, and Gang Yu.
\newblock Class-balanced grouping and sampling for point cloud 3d object
  detection.
\newblock \emph{arXiv preprint arXiv:1908.09492}, 2019.

\bibitem[Zhu et~al.(2020)Zhu, Su, Lu, Li, Wang, and Dai]{zhu2020deformable}
Xizhou Zhu, Weijie Su, Lewei Lu, Bin Li, Xiaogang Wang, and Jifeng Dai.
\newblock Deformable detr: Deformable transformers for end-to-end object
  detection.
\newblock In \emph{ICLR}, 2020.

\end{thebibliography}
}

\end{document}